\documentclass[letterpaper, 10 pt, conference]{ieeeconf}  

\IEEEoverridecommandlockouts                              

\usepackage{graphics} 
\usepackage{epsfig} 
\usepackage{mathptmx} 
\usepackage{times} 
\usepackage{amsmath} 
\usepackage{amssymb}  

\usepackage{multirow}
\usepackage{makecell}
\usepackage{booktabs}
\usepackage{rotating}
\usepackage{url}

\usepackage{graphicx,subcaption,ragged2e}

\setcellgapes{\tabcolsep}

\title{\LARGE \bf
Enhancing the Quality of 3D Lunar Maps Using JAXA’s Kaguya Imagery
}

\author{Yumi Iwashita$^{1}$, Haakon Moe$^{2}$, Yang Cheng$^{1}$, Adnan Ansar$^{1}$, Georgios Georgakis$^{1}$, Adrian Stoica$^{3}$, \\Kazuto Nakashima$^{4}$, Ryo Kurazume$^{4}$, Jim Torresen$^{2}$
\thanks{$^{1}$Y. Iwashita, Y, Cheng, A. Ansar, and G. Georgakis are with Jet Propulsion Laboratory, California Institute of Technology, Pasadena, CA, US
        {\tt\small yumi.iwashita@jpl.nasa.gov}}
\thanks{$^{2}$H. Moe and J. Torresen are with Department of Informatics, University of Oslo, Oslo, Norway}%
\thanks{$^{3}$A. Stoica is with LunaSol Space LLC}%
\thanks{$^{4}$K. Nakashima and R. Kurazume are with Department of Information Science and Electrical Engineering , Kyushu University, Fukuoka, Japan}%
}

\begin{document}

\maketitle
\thispagestyle{empty}
\pagestyle{empty}

\let\thefootnote\relax\footnotetext{Copyright © 2025. All rights reserved.}

\begin{abstract}
As global efforts to explore the Moon intensify, the need for high-quality 3D lunar maps becomes increasingly critical—particularly for long-distance missions such as NASA's Endurance mission concept, in which a rover aims to traverse 2,000 km across the South Pole–Aitken basin. Kaguya TC (Terrain Camera) images, though globally available at 10 m/pixel, suffer from altitude inaccuracies caused by stereo matching errors and JPEG-based compression artifacts. This paper presents a method to improve the quality of 3D maps generated from Kaguya TC images, focusing on mitigating the effects of compression-induced noise in disparity maps. We analyze the compression behavior of Kaguya TC imagery, and identify systematic disparity noise patterns, especially in darker regions. 
In this paper, we propose an approach to enhance 3D map quality by reducing residual noise in disparity images derived from compressed images. 
Our experimental results show that the proposed approach effectively reduces elevation noise, enhancing the safety and reliability of terrain data for future lunar missions.
\end{abstract}

\section{INTRODUCTION}

Space agencies around the world are competing to achieve successful missions on the Moon. China landed Chang’e on the far side of the Moon, while Japan successfully landed the SLIM mission—although it ran out of power due to the lander’s orientation relative to the Sun. The U.S. is pursuing the Artemis program to return humans to the Moon. The first private spacecraft, Intuitive Machines' Odysseus lander, successfully landed and operated for several days. NASA is also considering the Endurance mission to explore the lunar south pole.

The Endurance rover would cover 2,000 km in the South Pole-Aitken basin \cite{Endurance}. High-quality 3D maps are crucial for its safe and successful journey. LRO’s Lunar Orbiter Laser Altimeter (LOLA, 5~118 m/pixel) \cite{LOLA} and (JAXA) Kaguya’s Terrain Camera (TC, 10 m/pixel) \cite{TC} \cite{TCHaruyama} provide data for these maps, though existing ones have limitations. LRO's maps are high-resolution mainly in polar regions, but high-resolution mapping along Endurance's path requires new data collection. There's a concern as LRO may not be operational for the duration needed, due to constrains on original mission timeline. Kaguya maps, while high-resolution on the surface, have altitude inaccuracies from stereo matching and JPEG-based compression.

\begin{figure}
    \centering
    \includegraphics[width=6cm]{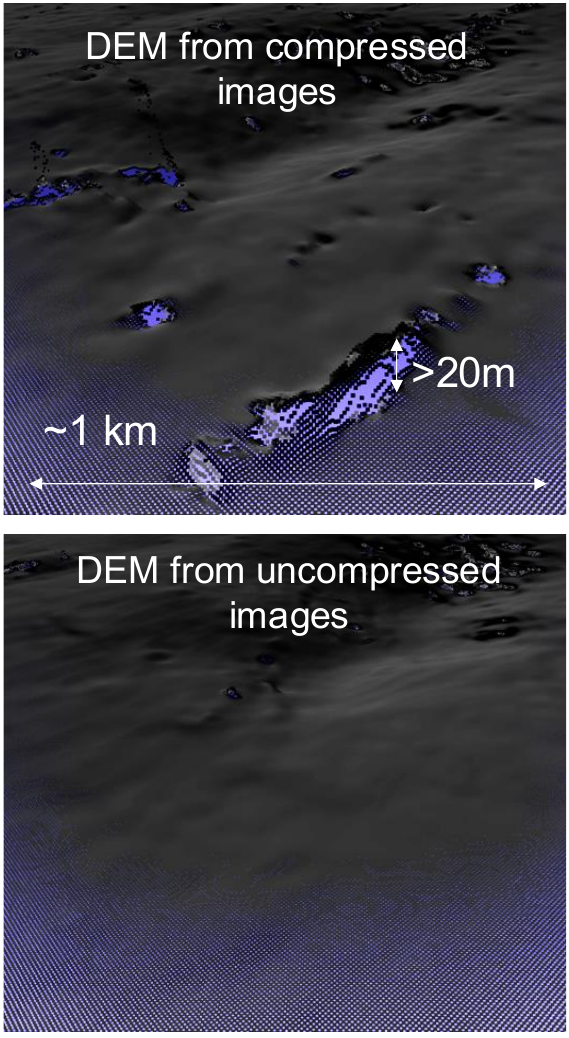}
    \caption{Illustrative example of elevation error (~20 m) in a 3D reconstruction derived from compressed stereo imagery (left). The DEM generated from uncompressed imagery of the same area (right) appears smooth.}
    \label{fig_20m}
\end{figure}

To address the first issue (i.e. altitude inaccuracies from stereo marching), we developed high-quality 3D lunar maps from Kaguya's 2D images \cite{LuNaMaps} \cite{YangKaguyaStereo}. Progress includes a Kaguya image selection system, TC camera model refinement, and enhanced stereo matching, resulting in less noisy 3D maps than JAXA’s. Despite better quality, challenges remain with Kaguya TC image compression affecting elevation accuracy. Analysis shows compression added about 20 meters of elevation noise. Figure \ref{fig_20m} generated from compressed and uncompressed images of the same area; the DEM (Digital Elevation Model) from the compressed image exhibits a 20-meter elevation error. This poses high risks for rover navigation due to potential underestimation of slope steepness and impassible areas, etc.




To address compression noise removal, both traditional signal processing techniques and more recent deep learning methods have been proposed. In signal processing approaches, Kamiya et al. developed a method to correct low-frequency components, offering improvements but often compromising high-frequency details \cite{2008ReductionOJ}. Singh et al. proposed mitigating blocking artifacts in JPEG images by modeling them as 2D step functions between neighboring blocks and applying human visual system-based metrics \cite{Singh}. 

Deep learning-based approaches for JPEG artifact reduction operate on fundamentally different principles, aiming for more comprehensive quality restoration. 
%
%
Recent models such as QGAC \cite{QGAC}, FBCNN \cite{fbcnn}, and DDRM-JPEG \cite{ddrm-jpeg} leverage techniques like Diffusion Models (DM) and advanced Convolutional Neural Network (CNN) to effectively reduce compression artifacts, particularly tested on ground-level imagery. Quantitative evaluations underscore their success, indicating that these state-of-the-art models can achieve over a 10\% improvement in standard image quality metrics compared to traditional decompression methods \cite{ddrm-jpeg}.


These approaches are generally not suitable for precise 3D reconstruction using satellite images, due to their coarse resolution compared to ground-level imagery. This coarse resolution often violates the continuity assumption commonly used in ground-based methods. While large craters may satisfy this assumption, our areas of interest are relatively flat regions—where landers are expected to touch down and rovers will operate—in order to ensure their safety.



Most Kaguya TC images are compressed, but approximately 5,000 remain uncompressed. These uncompressed images are geographically diverse, covering regions from the Moon’s South Pole to the North Pole. JAXA employs 32 different JPEG compression tables, selecting one for each image—information that is publicly available in each image’s metadata.
Our proposed approach leverages these compression tables. The initial idea was to generate 32 differently compressed versions of each uncompressed image and estimate the residuals between the compressed and uncompressed images using deep learning approaches. The goal was for the models to learn and capture patterns specific to JPEG compression artifacts. However, the residuals are random, making accurate estimation extremely challenging.

Through evaluation of disparity images generated by stereo matching, we observed that compressed stereo pairs introduced high-frequency noise in the disparity maps, whereas uncompressed pairs did not, as shown in Fig. \ref{fig_20m}. 
Thus in this paper, we propose an approach to enhance 3D map quality by improving disparity images derived from compressed inputs. We employ two deep learning approaches: IGEV++ for stereo matching \cite{xu2024igev} and a conditional diffusion model \cite{Palette} for post-processing disparity residuals. These methods are compared to determine which more effectively estimates residuals in the disparity images.

Post-processing methods for disparity images, such as PSMNet \cite{PSMNet} and RAFT-Stereo \cite{lipson2021raft}, have been proposed; however, they are designed to address typical matching errors, not compression-induced noise. Thus to the best of our knowledge, this is the first approach aimed at improving disparity images affected by compression-induced errors.

Our paper is organized as follows. Section II discusses JPEG compression in Kaguya TC images and describes our approach to improve disparity images. Section III presents the experimental results, and Section IV concludes the paper with a summary and directions for future work.

\section{JPEG COMPRESSION ON KAGUYA TC IMAGES AND IMPROVING DISPARITY IMAGES}
In this section, we first provide an overview of Kaguya TC images and examine how JPEG compression impacts the quality of disparity images, followed by our approach to enhance disparity accuracy.

\subsection{Kaguya TC images and JPEG compression}
\begin{figure}
    \centering
    \includegraphics[width=6cm]{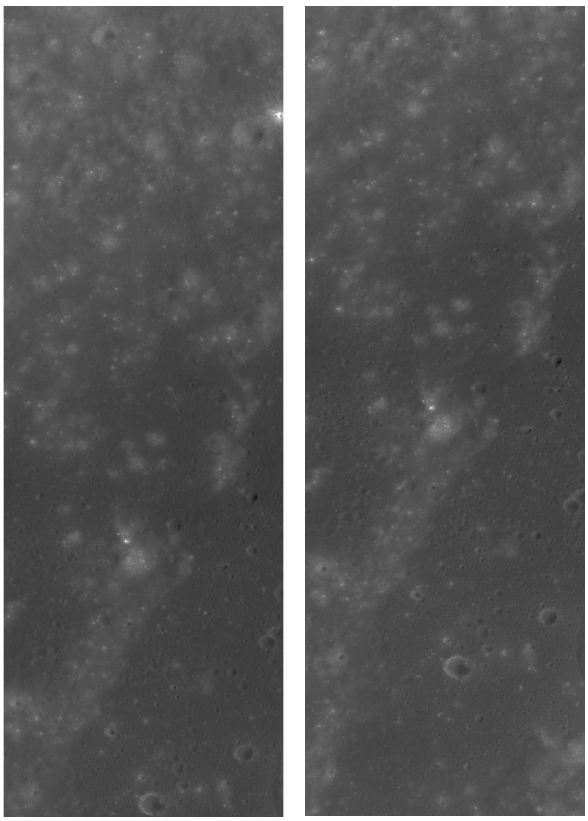}
    \caption{An example stereo pair (left: TC1W2B0$\_$01$\_$01685N188E0047, right: TC2W2B0$\_$01$\_$01685N185E0048).}
    \label{fig_stereo_pair}
\end{figure}

The Terrain Camera (TC) onboard the SELenological and ENgineering Explorer 'KAGUYA' (SELENE) spacecraft collected daytime stereo-pair imagery of the Moon from November 2007 to June 2009. The TC consists of two one-dimensional telescopes that captured images using a push-broom scanning method. The nominal image resolution is 10 m/pixel, based on Kaguya’s standard orbital altitude of 100 km. Each telescope has 4,096 pixels, with data typically acquired at 3,504 pixels, though 4,096 or 1,752 pixels were occasionally used.
TC operated in three observation modes: (i) stereoscopic mode at high solar elevation angles, (ii) monoscopic mode at lower solar elevation angles, and (iii) Spectral Profiler (SP) support mode. During the mission, Kaguya TC achieved stereo coverage of over 99$\%$ of the lunar surface \cite{Haruyama_12}. Images from the first observation mode were used in this study. An example stereo pair is shown in Fig. \ref{fig_stereo_pair}. 

\begin{table}[b]
    \centering
    \caption{JPEG compression table $SF008S\_A$. Left top is for low frequency and bottom right is for high frequency.}
    \begin{tabular}{c|c|c|c|c|c|c|c|c|}
    \multicolumn{1}{l}{} & \multicolumn{8}{c}{Low $\longleftrightarrow$ High (frequency)} \\\cline{2-9}
    \multirow{9}{*}{\rotatebox[origin=c]{90}{High $\longleftrightarrow$ Low}} 
    
    & 3 & 2 & 2 & 3 & 4 & 6 & 8 & 10 \\\cline{2-9}
    & 2 & 2 & 2 & 3 & 4 & 9 & 10 & 9 \\\cline{2-9}
    & 2 & 2 & 3 & 4 & 6 & 9 & 11 & 9 \\\cline{2-9}
    & 2 & 3 & 4 & 5 & 8 & 14 & 13 & 10 \\\cline{2-9}
    & 3 & 4 & 6 & 9 & 11 & 17 & 16 & 12 \\\cline{2-9}
    & 4 & 6 & 9 & 10 & 13 & 17 & 18 & 15 \\\cline{2-9}
    & 8 & 10 & 12 & 14 & 16 & 20 & 20 & 16 \\\cline{2-9}
    & 12 & 15 & 15 & 16 & 18 & 16 & 16 & 16 \\\cline{2-9}
    \end{tabular}
    \label{table_sf008s_a}
\end{table}

JAXA prepared 32 JPEG compression tables for Kaguya TC images, of which 10 were used during the mission. Among these, the most frequently used table is $SF008S\_A$, applied to approximately 56$\%$ of all images. Therefore, this paper focuses on $SF008S\_A$, as shown in Table \ref{table_sf008s_a}.
We do not delve into the theoretical details of JPEG compression in this paper; for a comprehensive explanation, please refer to \cite{JPEG}. Briefly, during the quantization step of JPEG compression, high-frequency components are suppressed, leading to residual loss—this is the primary source of compression-induced error.

In \cite{YangKaguyaStereo}, we demonstrated that our stereo matching algorithm can achieve sub-meter elevation accuracy when using uncompressed images. Based on the configuration of the TC cameras, the theoretical limit is a disparity error of 0.03 pixels, corresponding to a terrain relief error of approximately 0.54 meters along the y-direction (the spacecraft’s flight direction). In contrast, general stereo matching algorithms typically yield a disparity error of around 0.3 pixels, resulting in a terrain relief error of approximately 5.45 meters.

We investigated the impact of JPEG compression on the accuracy of disparity estimation. Our analysis revealed that disparity errors tend to increase in darker regions when using compressed images. To further evaluate this, we scaled the image DN (Digital Number, represented with 14-bit precision) values to various levels and applied JPEG compression noise using the $SF008S\_A$ table.

To obtain ground-truth disparity, we constructed stereo pairs by shifting a TC image by 97 pixels along the y-direction, resulting in a known ground-truth disparity of 97 pixels across the image. In our evaluation, we varied the image mean scale from 1.0 to 0.75, 0.5, 0.25, 0.1, and 0.05. For each scale, we computed disparity images from both uncompressed and compressed stereo pairs. The difference between the computed and ground-truth disparity was then used to calculate error statistics. The results are summarized in Table \ref{table_jpeg_noise}.
These results indicate that, even with varying image scaling, disparity estimation using uncompressed images maintains high accuracy, achieving a standard deviation of approximately 0.01 pixels. In contrast, performance degrades when using compressed images, particularly as DN values decrease. Since our goal is to achieve sub-meter accuracy in disparity estimation, this suggests the need to focus on improving performance in darker regions—specifically, areas with DN values below 388.

\begin{table*}
    \centering
    \caption{We varied the image mean scale from 1.0 to 0.75, 0.5, 0.25, 0.1, and 0.05. For each scale, we computed disparity images from both compressed and uncompressed stereo pairs. The difference between the computed and ground-truth disparity was then used to calculate error statistics.}
    \begin{tabular}{c|c|c|c|c|c|c|c|c|c|c}
    \multicolumn{3}{c|}{Image stats before compression} & \multicolumn{4}{c|}{Disparity image (compressed) [pixel]} & \multicolumn{4}{c}{Disparity image (uncompressed) [pixel]} \\\hline
    scale & mean [DN] & std [DN] & min & max & mean & std & min & max & mean & std \\\hline
    0.05 & 77.634 & 14.903 & 0.015 & 0.690 & 0.010 & \textbf{0.095} & 0.018 & 0.029 & -0.002 & \textbf{0.013} \\\hline
    0.10 & 155.268 & 29.806 & 0.014 & 0.778 & 0.001 & \textbf{0.060} & 0.004 & 0.025 & 0.000 & \textbf{0.013} \\\hline
    0.25 & 388.170 & 74.515 & 0.042 & 0.360 & -0.006 & \textbf{0.031} & 0.037 & 0.069 & -0.003 & \textbf{0.015} \\\hline
    0.50 & 776.340 & 149.030 & 0.048 & 0.322 & -0.005 & \textbf{0.023} & 0.007 & 0.080 & -0.000 & \textbf{0.016} \\\hline
    0.75 & 1164.510 & 223.545 & 0.047 & 0.181 & -0.002 & 0.016 & 0.046 & 0.046 & -0.002 & 0.014 \\\hline
    1.00 & 1552.680 & 298.060 & -0.031 & 0.208 & 0.002 & 0.017 & 0.063 & 0.056 & -0.002 & 0.015 \\\hline
    \end{tabular}
    \label{table_jpeg_noise}
\end{table*}

\subsection{Improve disparity images affected by compression-induced errors}
\begin{figure}
    \centering
    \includegraphics[width=6cm]{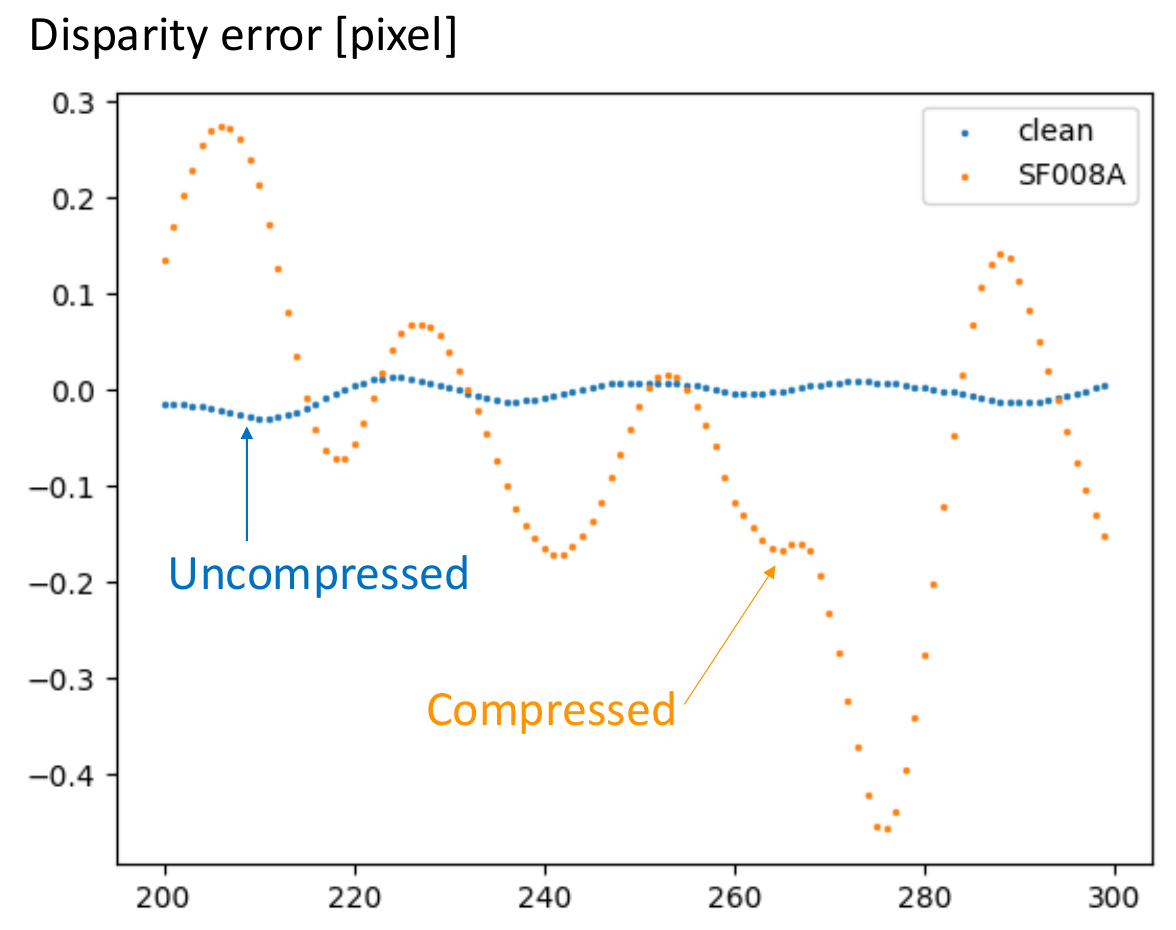}
    \caption{Illustrative comparison of disparity values from compressed (orange) and uncompressed (blue) images.}
    \label{fig_disp_noise}
\end{figure}

In this section, we present our approach to enhancing disparity images derived from compressed stereo pairs in darker regions. 
Figure \ref{fig_disp_noise} illustrates an example comparing disparity values along the y-axis. The blue and orange dots represent disparities from uncompressed and compressed stereo pairs, respectively. This observation led us to hypothesize that disparity noise from compressed images may exhibit patterns that can be mitigated, for instance, by removing high-frequency components.

As a proof of concept, we employ two deep learning methods: a conditional diffusion model (Palette) and a stereo matching model (IGEV++). 
Here is a brief explanation of both approaches. 
Palette \cite{Palette} is a conditional diffusion model developed for high-quality image-to-image translation tasks by learning to reverse a gradual noising process. It uses a U-Net architecture trained to predict noise added to the image at each timestep, and conditioning is done by concatenating the input with the noisy image at each step. 
%
IGEV++ \cite{xu2024igev} achieves state-of-the-art stereo matching by combining multi-scale feature extraction with multi-range geometry encoding volumes (MGEVs) for efficient disparity handling. It uses a ConvGRU-based operator to iteratively refine disparities and upsamples the final disparity map using context features.
IGEV++ is specifically designed for disparity estimation, and therefore is expected to perform well. However, diffusion models may also demonstrate competitive performance due to their inherently high expressive power.

Both models take a stereo pair of compressed images as input (or conditional data) and output the corresponding disparity map. Initially, we attempted to train the models directly using disparity maps from uncompressed images as ground truth.
However, training failed, likely due to the large range and skewed distribution of disparity values, which made diffusion model training challenging.
To address this, we analyzed the residuals between disparity maps generated from compressed and uncompressed image pairs and found that they follow a Gaussian distribution with zero mean and unit standard deviation. Based on this observation, we reformulated the problem to train the models to predict these residuals instead. To facilitate residual learning, we rectified the input stereo pairs of compressed images using the associated disparity maps.

\section{EXPERIMENTS}
In this section, we present experimental results using Kaguya TC images. To construct the dataset, we selected approximately 70 stereo pairs from the pool of 5,000 uncompressed images. The geographic coverage of these stereo pairs spans latitudes from -70 degree to 70 degree, with varying solar illumination conditions. 
For each stereo pair, we generated compressed versions and computed disparity maps for both the compressed and uncompressed image pairs.

As shown in the previous section (Table \ref{table_jpeg_noise}), the disparity error for uncompressed images is extremely small. Therefore, we treat the disparity maps derived from uncompressed images as ground truth in our evaluations.
%
Each original image has a resolution of 3208 × 4656 pixels, which we divided into smaller patches of 256 × 256 pixels. This resulted in 4745 patches, of which 90\% were used for training and 10\% for testing.


\begin{figure}
    \centering
\captionsetup[subfigure]{justification=Centering}

\begin{subfigure}{0.45\textwidth}
    \includegraphics[width=\textwidth]{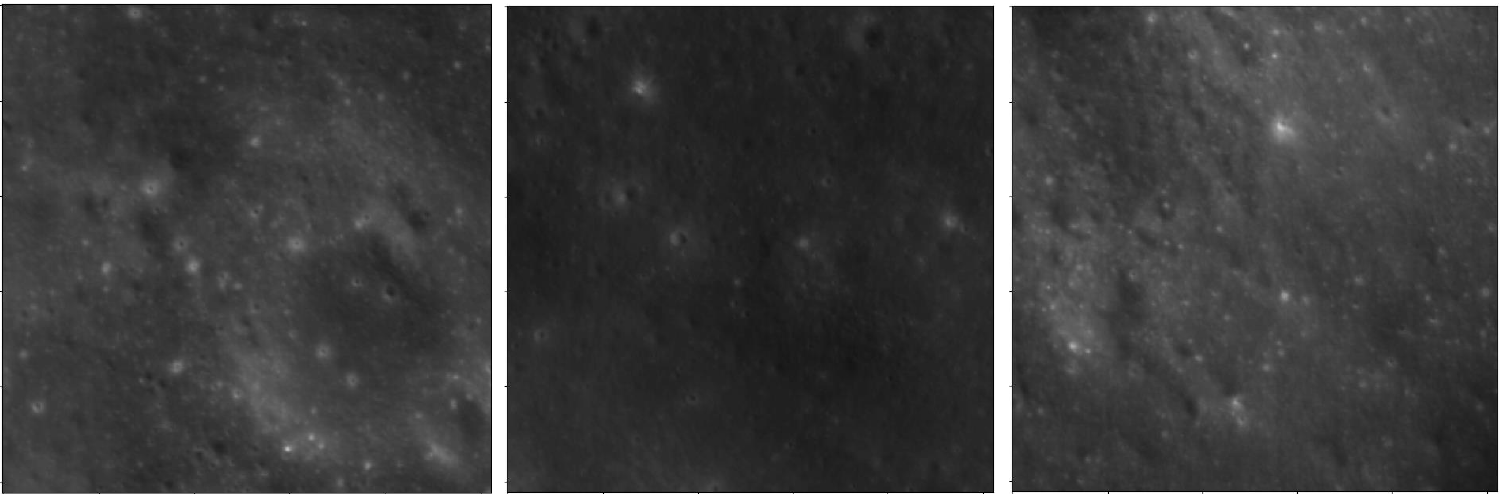}
    \caption{TC1 images}
\end{subfigure} 

\begin{subfigure}{0.45\textwidth}
    \includegraphics[width=\textwidth]{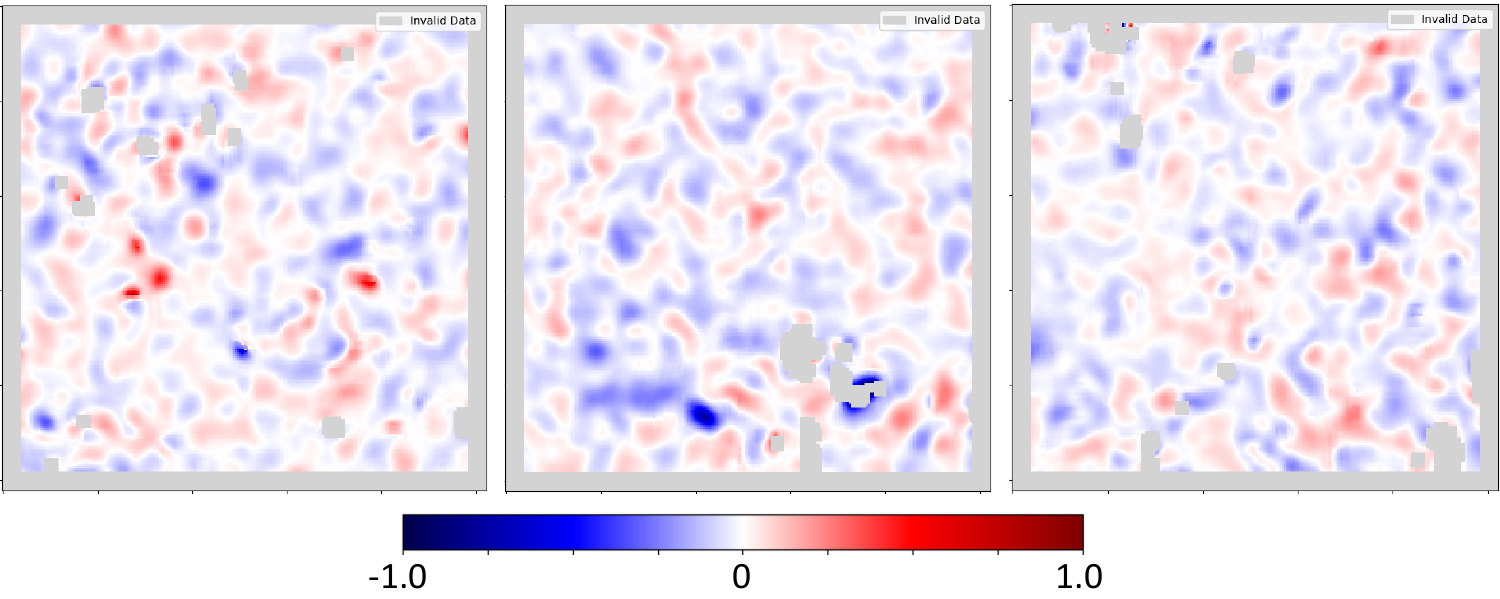}
    \caption{Ground truth disparity residuals}
\end{subfigure} 

\begin{subfigure}{0.45\textwidth}
    \includegraphics[width=\textwidth]{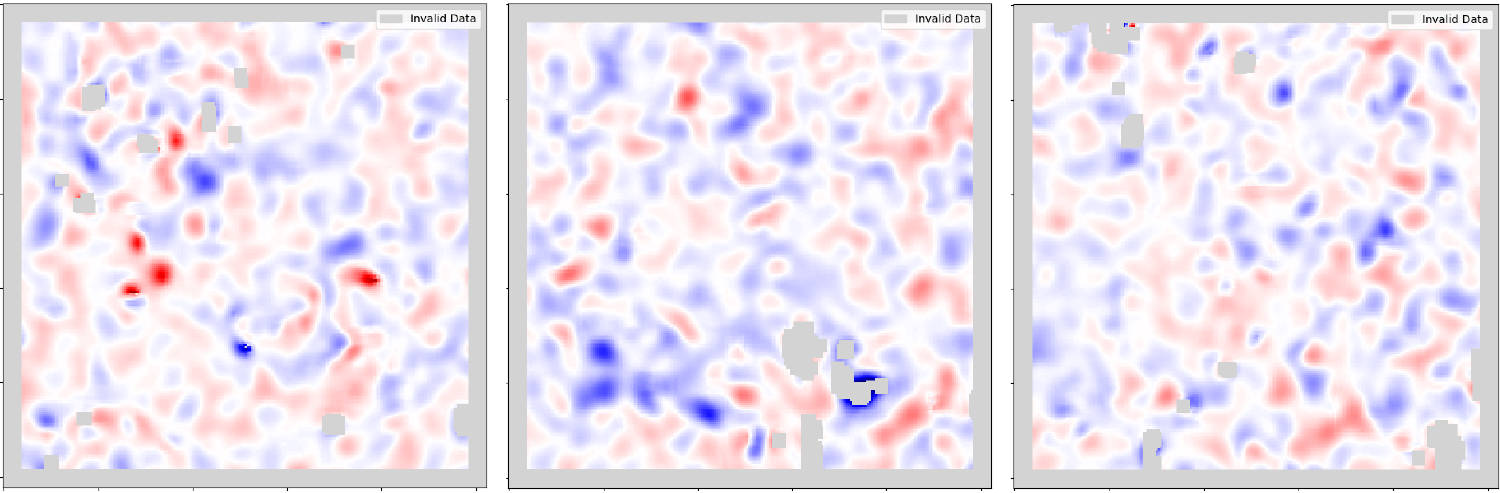}
    \caption{Estimation using Palette}
\end{subfigure} 

\begin{subfigure}{0.45\textwidth}
    \includegraphics[width=\textwidth]{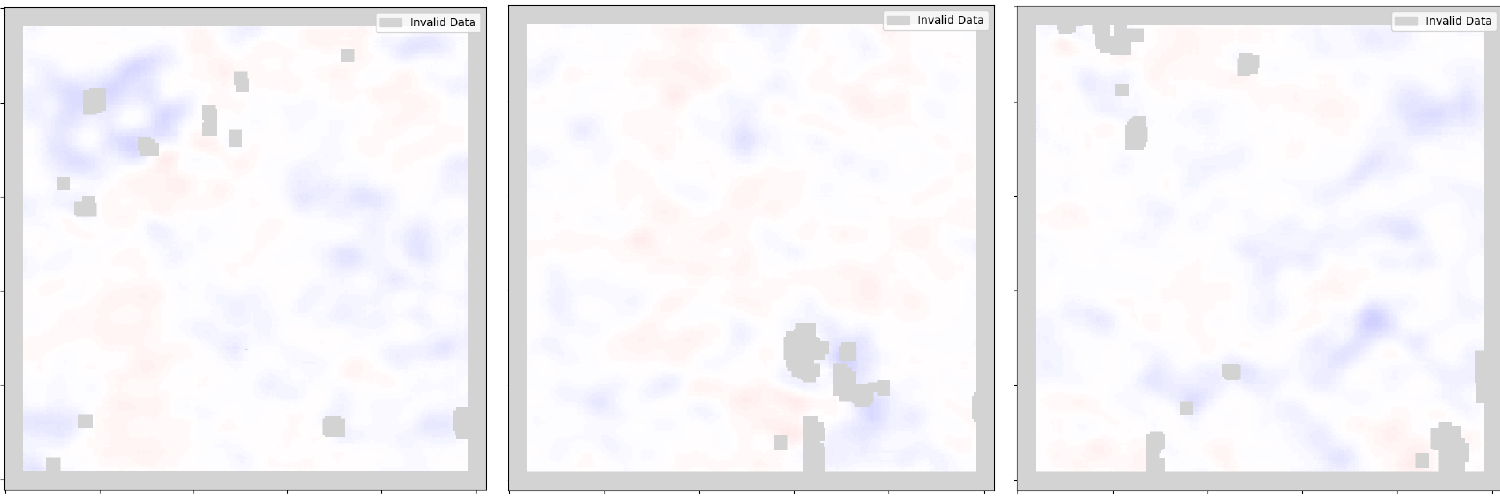}
    \caption{Estimation using IGEV++}
\end{subfigure} 

\begin{subfigure}{0.45\textwidth}
    \includegraphics[width=\textwidth]{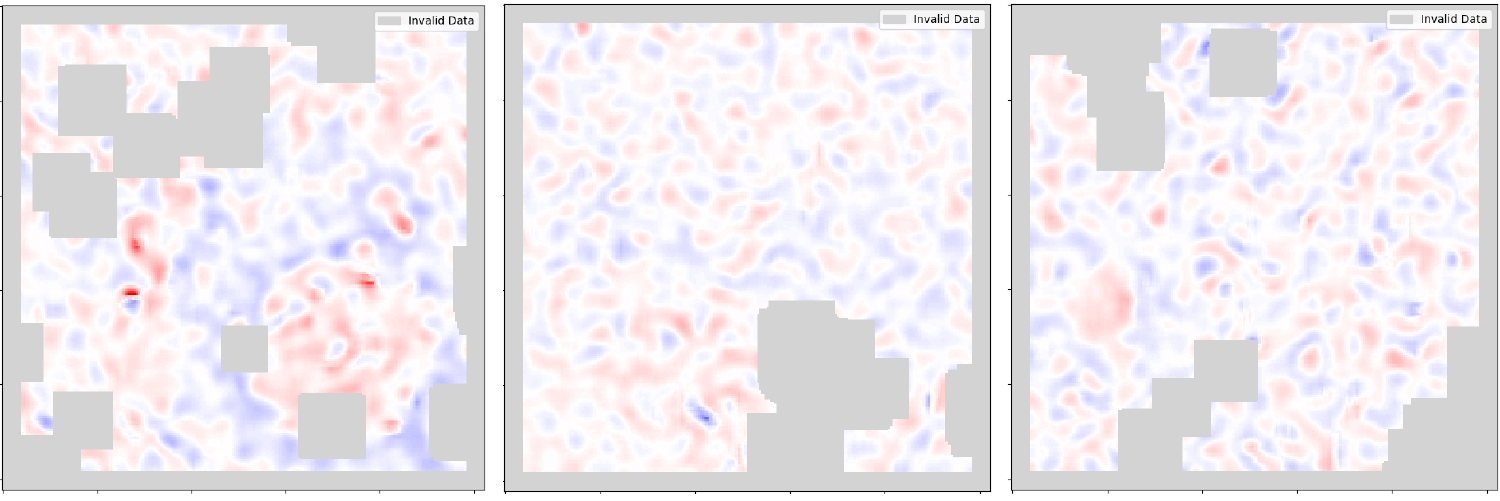}
    \caption{Estimation using LPF}
\end{subfigure} 

\caption{Example visualizations from the test dataset include: (a) TC1 images from three example rectified stereo pairs (Stereo pair IDs: 166, 615, and 918 from left to right); (b) ground truth disparity residuals corresponding to the stereo pairs in (a); and (c–e) estimated disparity residuals produced by Palette, IGEV++, and LPF}
    \label{fig_results}
\end{figure}

\begin{figure}
    \centering
\captionsetup[subfigure]{justification=Centering}

\begin{subfigure}{0.35\textwidth}
    \includegraphics[width=\textwidth]{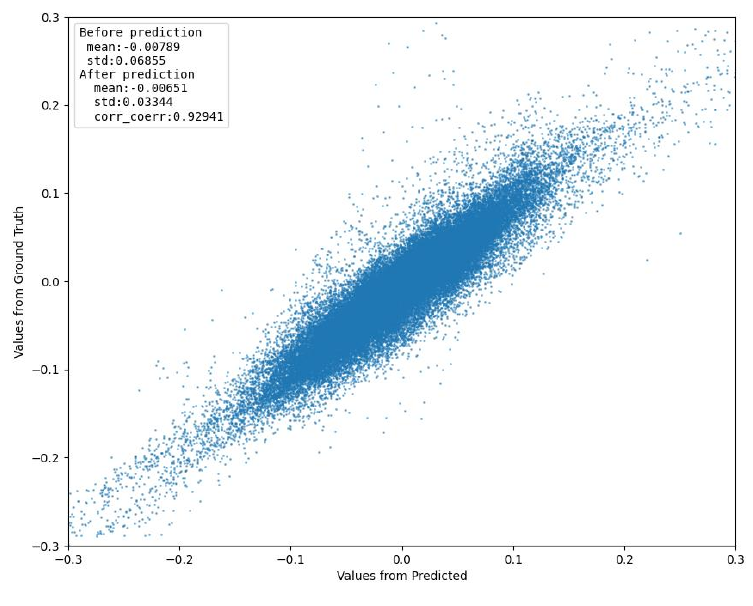}
    \caption{Palette}
\end{subfigure} 

\begin{subfigure}{0.35\textwidth}
    \includegraphics[width=\textwidth]{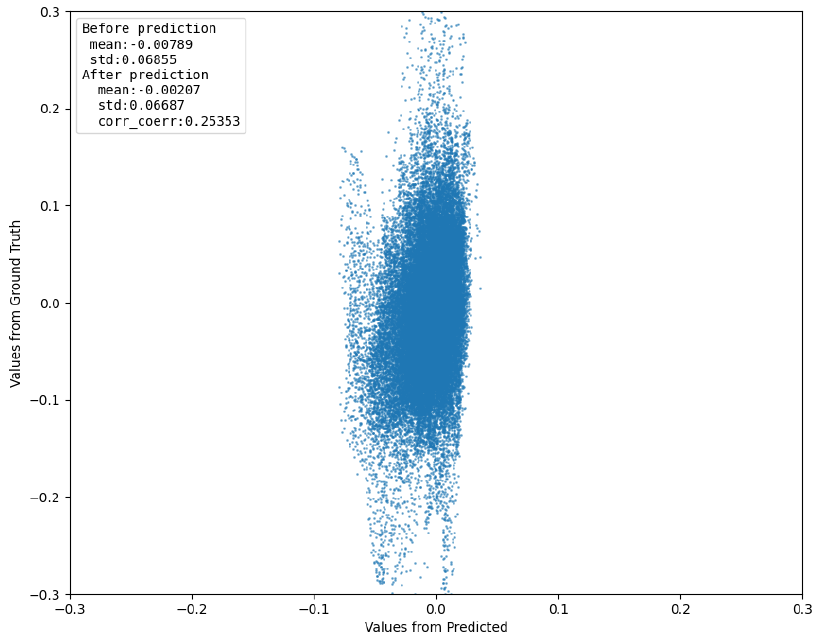}
    \caption{IGEV++}
\end{subfigure} 

\begin{subfigure}{0.35\textwidth}
    \includegraphics[width=\textwidth]{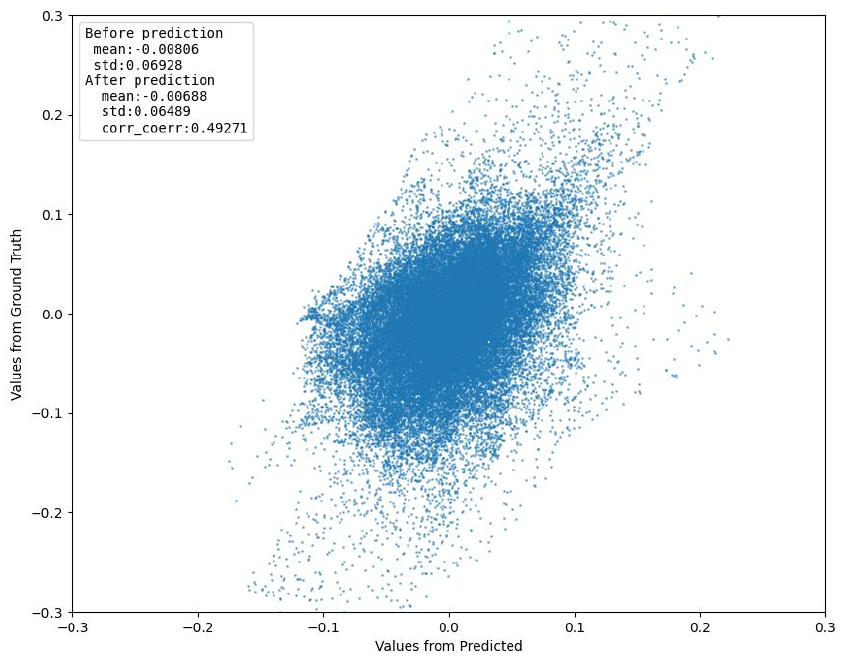}
    \caption{LPF}
\end{subfigure} 

    \caption{An example visualizations of 2D plots with the x-axis representing predictions and the y-axis representing ground truth (Stereo pair ID = 166).}
    \label{fig_2d_error_plots}
\end{figure}

\subsection{Comparison of Palette, IGEV++, and low pass filter}
We first trained a conditional diffusion model (Palette) and IGEV++ using the training data. For Palette, we used the Python package from \cite{Janspiry}, with key hyperparameters including a cosine scheduler with 2000 time steps and 1000 refinement steps. For IGEV++, we employed the official implementation \cite{IGEV++_code} with default settings. 
Additionally, we implemented a low-pass filter (LPF) to investigate the nature of noise in the disparity residuals. If the LPF produces good results, it suggests that the disparity errors mainly consist of high-frequency noise. However, if the LPF alone is insufficient, it indicates the presence of more complex noise components, which may require a data-driven approach (e.g., deep learning) for effective removal.

Figure \ref{fig_results} presents example visualizations from the test dataset, including TC1 images from example rectified stereo pairs, ground truth disparity residuals, estimated disparity residuals using Palette, IGEV++, and LPF. We picked up three examples from the test dataset, and stereo pair IDs are 166, 615, and 918. 
To implement LPF, we applied a Gaussian low-pass filter with a sigma of 3.0 to the original disparity images, then computed the residuals relative to the original disparities. 
These results indicate that Palette estimated the disparity residuals more accurately than IGEV++ and LPF.

Figure \ref{fig_2d_error_plots} shows 2D scatter plots with predictions on the x-axis and ground truth values on the y-axis for stereo pair ID 166.
These results indicate that Palette achieves the best performance, suggesting that the disparity noise consists not only of high-frequency components but also of other types of errors that Palette models effectively. 
IGEV++ shows the poorest performance, possibly because it is designed for larger disparity values and may have become trapped in a local minimum. 

We also computed the mean and standard deviation of the errors, the correlation coefficient, and the coefficient of determination between the estimated and ground truth disparity residuals for four methods—no noise estimation (i.e., the original disparity map from compressed images), Palette, IGEV++, and LPF—as summarized in Table \ref{table_exp} for the three example stereo pairs (IDs: 166, 615, and 918). 
The results show that the original disparity maps derived from compressed images have a standard deviation exceeding 0.06 pixels and a mean error of approximately 0.01 pixels, corresponding to an elevation error of about 1.26 meters. In contrast, the conditional diffusion model improves estimation accuracy, achieving a best-case standard deviation of 0.033 pixels and a mean error of 0.007 pixels, resulting in an elevation error of 0.54 meters. These findings demonstrate the feasibility of our proposed approach for enhancing disparity maps degraded by compression-induced noise.

Table \ref{table_exp_all} presents the correlation coefficient and the coefficient of determination for the overall evaluation using Palette, IGEV++, and LPF.

\begin{table*}
    \centering
    \caption{Evaluation of three example image sets using: (i) no noise estimation (i.e., the original disparity map generated from compressed images), (ii) Palette, (iii) IGEV++, and (iv) LPF. $r$ denotes the correlation coefficient, and $r^2$ denotes the coefficient of determination ($R^2$ score). $r$ and $r^2$ values are not available for (i) no estimation, since the residual of the disparity map is all 0. SD stands for standard deviation.}
    \begin{tabular}{c|c|c|c|c|c|c|c|c|c|c|c|c|c|c|c|c}
    ID & \multicolumn{4}{c|}{(i) no estimation } & \multicolumn{4}{c|}{(ii) Palette} & \multicolumn{4}{c|}{(iii) IGEV++} & \multicolumn{4}{c}{(iv) LPF} \\\cline{2-17}
     & mean & SD & $r$ & $r^2$ & mean & SD & $r$ & $r^2$ & mean & SD & $r$ & $r^2$ & mean & SD & $r$ & $r^2$ \\\hline    
    166 & 0.008 & 0.069 & N/A & N/A & 0.007 & 0.033 & 0.929 & 0.844 & -0.002 & 0.067 & 0.253 & 0.063 & 0.006 & 0.065 & 0.493 & 0.186 \\\hline
    615 & 0.015 & 0.072 & N/A & N/A & 0.002 & 0.056 & 0.728 & 0.500 & 0.015 & 0.071 & 0.215 & -0.009 & 0.014 & 0.068 & 0.440 & 0.130 \\\hline
    918 & 0.013 & 0.063 & N/A & N/A & 0.006 & 0.040 & 0.866 & 0.715 & -0.007 & 0.062 & 0.167 & 0.015 & 0.013 & 0.064 & 0.454 & 0.174 \\\hline
    \end{tabular}
    \label{table_exp}
\end{table*}

\begin{table}
    \centering
    \caption{Overall evaluation of the following methods: (i) using Palette, (ii) using IGEV++, and (iii) using LPF. $r$ denotes the correlation coefficient, and $r^2$ denotes the coefficient of determination.}
    \begin{tabular}{c|c|c}
     & $r$ & $r^2$ \\\hline
    (i) Palette & 0.778 & 0.487 \\\hline
    (ii) IGEV++ & 0.166 & 0.017 \\\hline
    (iii) LPF & 0.463 & 0.193 \\\hline
    \end{tabular}
    \label{table_exp_all}
\end{table}

\section{CONCLUSIONS AND FUTURE WORK}
We proposed the first approach specifically targeting the improvement of disparity images affected by compression-induced noise in Kaguya TC images. By leveraging deep learning models, we demonstrate that it is possible to enhance 3D map quality despite the challenges posed by JPEG compression artifacts. 

Our future work includes evaluating our approach across a range of mean DN values, as current testing has been limited to a mean DN of 200. We also plan to assess the effectiveness of our method using the remaining nine JPEG compression tables provided by JAXA.



\section*{ACKNOWLEDGMENT}
The research described in this paper was carried out at the Jet Propulsion Laboratory, California Institute of Technology, under a contract with the National Aeronautics and Space Administration. Government sponsorship acknowledged. 
This work was partly supported by the Research Council of Norway and Diku project Collaboration on Intelligent Machines (COINMAC-2) project, under grant agreement 309869.
We would like to express our sincere gratitude to Dr. Junichi Haruyama of JAXA for providing the JPEG compression table.


\begin{thebibliography}{99}

\bibitem{Endurance}
Mission Concept Study Report for the 2023–2032 Planetary Science and Astrobiology Decadal Survey, \url{https://science.nasa.gov/wp-content/uploads/2023/11/endurance-spa-traverse-and-sample-return.pdf}

\bibitem{LOLA}
Lunar Orbiter Laser Altimeter, \url{https://science.nasa.gov/mission/lro/lola/}

\bibitem{TC}
Kaguya Terrain Camera, \textit{https://www.kaguya.jaxa.jp/en/equipment/tc$\_$e.htm}

\bibitem{TCHaruyama}
J. Haruyama, M. Ohtake, T. Matsunaga, T. Morota, Y. Yokota, C. Honda, N. Hirata, H. Demura, A. Iwasaki, R. Nakamura, S. Kodama,
Planned radiometrically calibrated and geometrically corrected products of lunar high-resolution Terrain Camera on SELENE,
Advances in Space Research,
Volume 42, Issue 2,
Pages 310-316, 2008.

\bibitem{LuNaMaps}
C. Restrepo, N. Petro, M. Barker, E. Mazarico, S. Scheidt, J. Richardson, S. Bertone, C. Gnam, A. Liounis, R. Beyer, Y. Cheng, A. Ansar, C. Mauceri, Y. Iwashita, Z. Morgan, LuNaMaps Project, https://ntrs.nasa.gov/citations/20240012328.

\bibitem{YangKaguyaStereo}
Y. Cheng, A. Ansar, Y. Iwashita, KAGUYA Terrain CAmera DEM Improvement, 4th Space Imaging Workshop, 2024. 


\bibitem{Palette}
Chitwan Saharia, et al., “Palette: Image-to-Image Diffusion Models”, SIGGRAPH, 2022.

\bibitem{2008ReductionOJ}
U. Kamiya, Reduction of JPEG noise from the ALOS PRISM products, https://api.semanticscholar.org/CorpusID:189859444, 2008.

\bibitem{Singh}
S. Singh, V. Kuma, H. Verma, Reduction of blocking artifacts in JPEG compressed images, Digital Signal Processing, 2007. 

\bibitem{PSMNet}
J. Chang, Y. Chen, Pyramid Stereo Matching Network, CVPR 2018.

\bibitem{lipson2021raft}
L. Lipson, Z. Teed, J. Deng, RAFT-Stereo: Multilevel Recurrent Field Transforms for Stereo Matching, 3DV 2021


\bibitem{JPEG}
G. K. Wallace, The JPEG still picture compression standard, IEEE Trans. on Consumer Electronics, vol. 38, no. 1, pp. xviii-xxxiv, Feb. 1992, doi: 10.1109/30.125072.

\bibitem{Haruyama_12}
J. Haruyama, et al. Lunar Global Digital Terrain Model Dataset Produced from SELENE (KAGUYA) Terrain Camera Stereo Observations, 43rd Lunar and Planetary Science Conference, 2012.

\bibitem{QGAC}
M. Ehrlich, L. Davis, S.-N. Lim, and A. Shrivastava, 
\textit{Quantization Guided JPEG Artifact Correction}, 
arXiv preprint arXiv:2004.09320, 2020. Available: \url{https://arxiv.org/abs/2004.09320}

\bibitem{ddrm-jpeg}
B. Kawar, J. Song, S. Ermon, and M. Elad, 
\textit{JPEG Artifact Correction using Denoising Diffusion Restoration Models}, 
arXiv preprint arXiv:2209.11888, 2022. Available: \url{https://arxiv.org/abs/2209.11888}

\bibitem{fbcnn}
J. Jiang, K. Zhang, and R. Timofte, 
\textit{Towards Flexible Blind JPEG Artifacts Removal}, 
arXiv preprint arXiv:2109.14573, 2021. Available: \url{https://arxiv.org/abs/2109.14573}

\bibitem{xu2024igev}
Gangwei Xu, Xianqi Wang, Zhaoxing Zhang, Junda Cheng, Chunyuan Liao, and Xin Yang.
\newblock \textit{IGEV++: Iterative Multi-range Geometry Encoding Volumes for Stereo Matching}.
\newblock 2024.
\newblock arXiv:2409.00638 [cs.CV]. Available at: \url{https://arxiv.org/abs/2409.00638}.

\bibitem{Janspiry}
https://github.com/Janspiry/Palette-Image-to-Image-Diffusion-Models

\bibitem{IGEV++_code}
https://github.com/gangweix/IGEV-plusplus



\end{thebibliography}
\end{document}